\pdfoutput=1

\documentclass[11pt]{article}

\usepackage[final]{acl}

\usepackage{times}
\usepackage{latexsym}

\usepackage[T1]{fontenc}

\usepackage[utf8]{inputenc}

\usepackage{microtype}

\usepackage{inconsolata}

\usepackage{amsmath}
\usepackage{amssymb}

\usepackage{algorithm}
\usepackage{algpseudocode}
\usepackage{xcolor}
\usepackage{graphicx}
\usepackage{wrapfig}
\usepackage{subcaption}

\usepackage{verbatim}
\usepackage{tabularx}

\newcommand{\sg}[1]{\textcolor{blue}{#1}}
\newcommand{\pl}[1]{\textcolor{red}{#1}}
\newcommand{\np}[1]{\textcolor{teal}{#1}}
\newcommand{\agt}[1]{\textcolor{violet}{#1}}
\newcommand{\pp}[1]{\textcolor{orange}{#1}}
\newcommand{\thm}[1]{\textcolor{blue}{#1}}
\newcommand{\loc}[1]{\textcolor{red}{#1}}

\newcommand{\ul}[1]{\underline{#1}}

\title{Tracking linguistic information in transformer-based sentence embeddings \\ through targeted sparsification}

\author{
  \textbf{Vivi Nastase\textsuperscript{1}} \and
  \textbf{Paola Merlo\textsuperscript{1,2}}
\\
  \textsuperscript{1}Idiap Research Institute, Martigny, Switzerland \\
  \textsuperscript{2}University of Geneva, Swizerland
  \\
\texttt{vivi.a.nastase@gmail.com, Paola.Merlo@unige.ch}
}

\begin{document}
\maketitle
\begin{abstract}

Analyses of transformer-based models have shown that they encode a variety of linguistic information from their textual input. While these analyses have shed a light on the relation between linguistic information on one side, and internal architecture and parameters on the other, a question remains unanswered: how is this linguistic information reflected in sentence embeddings? Using datasets consisting of sentences with known structure, we test to what degree information about chunks (in particular noun, verb or prepositional phrases), such as grammatical number, or semantic role, can be localized in sentence embeddings. Our results show that such information is not distributed over the entire sentence embedding, but rather it is encoded in specific regions. Understanding how the information from an input text is compressed into sentence embeddings helps understand current transformer models and help build future explainable neural models.

\end{abstract}



\section{Introduction}

In the quest for understanding transformer-based models, much work has been dedicated to uncover what kind of information is encoded in the model's various layers and parameters.
These analyses have provided several enlightening insights: (i)  different types of linguistic information -- e.g. parts of speech, syntactic structure, named entities -- are selectively more evident at different layers of the model \citep{tenney-etal-2019-bert,rogers-etal-2020-primer}, (ii) subnetworks can be identified that seem to encode particular linguistic functionalities \citep{csordas2021neural}, and (iii) fine-tuning for specific tasks can be focused on very small subsets of parameters, on different parts of a model's layers \citep{panigrahi2023task}. While these analyses and probes have focused on the insides of the models, mostly their parameters and layers, testing their impact is usually done by using the output of the model, namely token or sentence embeddings, to solve specific tasks. The link between the inside of the model and its outputs is usually not explicitly investigated. 

We ask several facets of  this question here:   
how are the internally-detected information types and structures reflected in the model's output? And how are arbitrarily long and complex sentences encoded systematically in a fixed-sized vector? 

Understanding what kind of information the sentence embeddings encode, and how, has multiple benefits: (i) it connects internal changes in the model parameters and structure with changes in its outputs; (ii) it contributes to verifying the robustness of models and whether or not they rely on shallow or accidental regularities in the data; (iii) it narrows down the field of search when a language model produces wrong outputs, and (iv) it helps maximize the use of training data for developing more robust models from smaller textual resources.

Transformer-based models usually use a token-focused learning objective, and have a weaker supervision signal at the sentence level -- e.g. a next sentence prediction \citep{devlin-etal-2018-bert}, or sentence order information \citep{lan-etal-2020-albert}. Despite this focus, high performance in a variety of tasks (using raw or fine-tuned sentence embeddings) as well as direct probing shows that sentence representations encode a variety of linguistic information \cite{conneau-etal-2018-cram}. 
On the other hand,  direct exploration of BERT sentence embeddings has also shown that they contain mostly shallow information, related to sentence length and lexical variation, and that many of their dimensions are correlated, indicating that either information is redundantly encoded, or that not all dimensions encode useful information \cite{nikolaev-pado-2023-universe}. Some of this preexisting work assumes that sentence embeddings encode information in an overt manner, for example,  each principal component dimension is responsible for encoding some type of information. 

We adopt the different view that information in sentence embeddings may be encoded in merged layers, in a manner similar to audio signals being composed of overlapping signals of different frequencies. We hypothesize that each such layer may encode different types of information. We aim to test this hypothesis and check (i) whether we can separate such layers, and (ii) investigate whether information about specific chunks in a sentence --noun,verb, or prepositional phrases-- is encoded in different layers and parts of a sentence embedding.

We perform our investigation in an environment with data focused on specific grammatical phenomena, while displaying lexical, structural and semantic variation, and a previously developed system that has been shown to detect  the targeted phenomena well \cite{nastase2024identifiable}. The system is a variational encoder-decoder, with an encoder that compresses the information in the input into a very low-dimensional latent vector. \citet{nastase2024identifiable}~ have shown that the sentence embeddings, and their compressed representations on the latent layer, encode information about chunks -- noun, verb, prepositional phrases -- and their linguistic properties. 


The current study investigates the general hypothesis indicated above by specifically exploring two new research questions in this setting:
\begin{enumerate}
    \item Whether a targeted sparsification of the system 
    maintains a high performance on the task, indicating that information about chunks in the sentence is localizable. 
    \item Contingent on the answer to the first question, we trace back the signal from the latent layer to the input sentence embeddings, and analyze how specific differences in chunk properties -- different number of chunks, or chunks that differ from each other only on one property (e.g. grammatical number) -- are localized and reflected in the sentence embeddings.
\end{enumerate}

The code and data are available at \url{https://github.com/CLCL-Geneva/BLM-SNFDisentangling}.

\section{Related work}

\paragraph{Sentence embeddings} Transformer models induce contextual token embeddings by passing the embedding vectors through successive layers using multi-head attention that allows for tokens to influence each other's representation at each successive step \cite{vaswani2017attention}. The model focuses on the token embeddings, as the tokens expected on the output layer provide the training signal. There are numerous variations on the BERT \citep{devlin-etal-2018-bert} transformer model\footnote{\url{https://huggingface.co/docs/transformers/en/model_summary}}, that vary in the way the models are trained \cite{liu2019roberta}, how they combine (or not) the positional and token embeddings \citep{he-etal-2020-deberta}, how the input is presented to the model \citep{liu2019roberta,clark2020electra}. With regards to the sentence-level supervision signal, BERT \citep{devlin-etal-2018-bert} uses the next sentence prediction objective, ALBERT \citep{lan-etal-2020-albert}, aiming to improve coherence, uses sentence order prediction. It is more common to further train or fine-tune a pre-trained model to produce sentence embeddings fitting specific tasks, such as story continuation \citep{ippolito-etal-2020-toward} or sentence similarity \citep{reimers-gurevych-2019-sentence}.

Electra \cite{clark2020electra} does not have a sentence-level objective, but it relies on replaced token detection, which relies on the sentence context to determine whether a (number of) token(s) in the given sentence were replaced by a generator sample. This leads to sentence embeddings that perform well on tasks such as Question Answering, or detecting verb classes \citep{yi-etal-2022-probing}.

\paragraph{Probing embeddings and models for linguistic information}

Most work investigating the kind of knowledge captured by transformer-based models have focused on analysing the architecture of the model \citep{tenney-etal-2019-learn-from-context,rogers-etal-2020-primer} to determine the localization and flow of information through the model's layers. There is also much work on analyzing the induced token embeddings to determine what kind of linguistic information they encode, such as sentence structure \citep{hewitt-manning-2019-structural}, predicate argument structure \citep{conia-etal-2022-semantic}, subjecthood and objecthood \citep{papadimitriou-etal-2021-deep}, among others. Testing whether sentence representation contain specific types of linguistic information has been done using task (or information)-specific classifiers  \citep{adi-etal-2017-finegrained,conneau-etal-2018-cram,goldberg2019,wilson-etal-2023-abstract}. \citet{opitz-frank-2022-sbert} aim to map subsets of dimensions of fine-tuned sentence embeddings to semantic features.

\paragraph{Sparsification} Deep learning models have billions of parameters. This makes them not only incomprehensible, but also expensive to train. The lottery ticket hypothesis \citep{frankle-carbin-2018-lottery} posits that large networks can be reduced to subnetworks that encode efficiently the functionality of the entire network. Detecting functional subnetworks can be done \textit{a posteriori}, over a pre-learned network to investigate the functionality of detected subnetworks \citep{csordas2021neural}, the potential compositionality of the learned model \citep{lepori2023break}, or where task-specific skills are encoded in a fine-tuned model \citep{panigrahi2023task}.

Instead of learning a sparse network over a prelearned model, \citet{cao-etal-2021-low} use a pruning-based approach to finding subnetworks in a pretrained model that performs some linguistic task. Pruning can be done at several levels of granularity: weights, neurons, layers. Their analyses confirm previous investigations of the types of information encoded in different layers of a transformer \citep{conneau-etal-2018-cram}. \citet{conmy-etal-2023-circuitdiscovery} introduce the Automatic Circuit DisCovery (ACDC) algorithm, which adapts subnetwork probing and head importance score for pruning to discover circuits that implement specific linguistic functions.

Sparsification can also be achieved using $L_0$ regularization, as the pruning would be done directly during training by encouraging weights to become exactly zero. \citet{louizos-etal-2018-l0regularization,savarese2020-sparsification}, among others, implement solutions to the issue that $L_0$ regularization is non-differentiable, and test it on image classification.




\vspace{2mm}

\noindent The cited work focuses on the parameters of the model, and sparsification approaches aiming to detect the subnetworks to which specific skills or linguistic information can be ascribed. Our focus, instead,  is the output of transformer-based models, in particular sentence embeddings, which we investigate using targeted sparsification. 

\section{Approach overview}

We investigate whether we can identify specific sentence properties in sentence embeddings. \citet{nastase2024identifiable} have shown that using an encoder-decoder architecture, sentence embeddings can be compressed into a latent representation that preserves information about chunks in a sentence, and their properties necessary to solve a specific linguistic task. 


We first test whether we can sparsify this architecture in a targeted manner, such that each region of the sentence embedding contributes a signal to only one unit of the latent layer. This allows us to isolate different parts of the sentence embedding. 

After establishing that sparsification does not lead to a dramatic drop in performance, we trace back the signal from the latent layer to the sentence embeddings, and test whether we can localize information about how different numbers of chunks, or chunks with different properties, are encoded.

In the final step, we use the sparse encoder-decoder sentence compression system as the first in a two-layer system used to solve language tasks -- called Blackbird Language Matrices \cite{merlo2023} -- that require chunk and chunk properties information. The first layer will compress each sentence into a very small latent vector, and this representation is then used on the second layer to solve a pattern detection problem that relies on information about chunks in a sentence and their pattern across a sequence of sentences. 

\section{Data}
\label{sec:data}

We use two data types: (i) a dataset of sentences with known chunk structure and chunk properties, (ii) two datasets representing two multiple-choice problems, whose solution requires understanding the chunk structure and chunk properties of the sentences in each instance.

\begin{figure*}[h]
\begin{minipage}{0.45\textwidth}
\small
\begin{tabular}{llll} 
\multicolumn{4}{l}{BLM agreement problem} \\
\hline
\multicolumn{4}{c}{\sc Context Template}\\
\hline
\sg{NP-sg}& \sg{PP1-sg}& & \sg{VP-sg}  \\
\pl{NP-pl} & \sg{PP1-sg}& & \pl{VP-pl}  \\
\sg{NP-sg}& \pl{PP1-pl} & & \sg{VP-sg}  \\
\pl{NP-pl} & \pl{PP1-pl} & & \pl{VP-pl}  \\
\sg{NP-sg}& \sg{PP1-sg}& \sg{PP2-sg} &\sg{VP-sg}  \\
\pl{NP-pl} & \sg{PP1-sg}  & \sg{PP2-sg} &            \pl{VP-pl}  \\
\sg{NP-sg}   & \pl{PP1-pl} & \sg{PP2-sg} &             \sg{VP-sg}  \\
\end{tabular}

  \begin{tabular}{ll} \hline
 \multicolumn{2}{c}{{\sc Answer set} } \\ \hline
NP-sg PP1-sg et NP2 VP-sg & Coord  \\ 
{\bf NP-pl PP1-pl NP2-sg VP-pl } & correct \\
NP-sg PP1-sg VP-sg & WNA\\
NP-pl PP1-pl NP2-pl VP-sg & AE\_V \\
NP-pl PP1-sg NP2-pl VP-sg & AE\_N1 \\
NP-pl PP1-pl NP2-sg VP-sg & AE\_N2 \\
NP-pl PP1-sg PP1-sg VP-pl & WN1 \\
NP-pl PP1-pl PP2-pl VP-pl & WN2 \\
 \hline
 \end{tabular}
\end{minipage}
\hfill
\begin{minipage}{0.55\textwidth}
\small
\setlength{\tabcolsep}{3pt} 
\begin{tabular}{llll} 
\multicolumn{4}{l}{BLM verb alternation problem} \\
\hline
\multicolumn{4}{c}{\sc Context Template}\\
\hline
\np{NP-\agt{Agent}} & Verb & \np{NP-}\loc{Loc} & \pp{PP-}\thm{Theme} \\
\np{NP-\thm{Theme}} & VerbPass & \pp{PP-}\agt{Agent} \\
\np{NP-\thm{Theme}} & VerbPass & \pp{PP-}\loc{Loc} & \pp{PP-}\agt{Agent}\\
\np{NP-\thm{Theme}} & VerbPass & \pp{PP-}\loc{Loc}  \\
\np{NP-\loc{Loc}} & VerbPass & \pp{PP-}\agt{Agent} \\
\np{NP-\loc{Loc}} & VerbPass & \pp{PP-}\thm{Theme} & \pp{PP-}\agt{Agent} \\
\np{NP-\loc{Loc}} & VerbPass & \pp{PP-}\thm{Theme} \\
\end{tabular}
\begin{tabular}{ll} \hline
\multicolumn{2}{c}{\sc Answer set}\\ \hline
\bf{NP-Agent Verb NP-Theme PP-Loc} & \textsc{Correct} \\ 
NP-Agent *VerbPass NP-Theme PP-Loc & \textsc{AgentAct}\\
NP-Agent Verb NP-Theme *NP-Loc & \textsc{Alt1}\\
NP-Agent Verb *PP-Theme PP-Loc & \textsc{Alt2}\\
NP-Agent Verb *[NP-Theme PP-Loc] & \textsc{NoEmb}\\
NP-Agent Verb NP-Theme *PP-Loc & \textsc{LexPrep}\\
*NP-Theme Verb NP-Agent PP-Loc & \textsc{SSM1}\\
*NP-Loc Verb NP-Agent PP-Theme &\textsc{SSM2}\\
*NP-Theme Verb NP-Loc PP-Agent & \textsc{AASSM}\\
\hline
 \end{tabular}
\label{ALT-ATL}
\end{minipage}
\caption{Structure of two BLM problems, in terms of chunks in sentences and sequence structure.}
\label{fig:BLM_structure}
\end{figure*}

\subsection{A dataset of sentences}
\label{sec:sent_descr}

We start with an artificially-created set of sentences built from noun, prepositional and verb phrases. Each sentence has one of the following structures: {\rm NP [PP$_1$ [PP$_2$]] VP }, where the parentheses surround optional structure. Each chunk can have singular or plural form, with agreement between the first NP (the subject) and the VP. This leads to 14'336 sentences with one of 14 patterns.

The dataset consists of ordered pairs of one input sentence and $N$ (=7) output sentences, extracted from the set described above. Only one of the output sentences has the same chunk pattern as the input sentence, and is considered as the correct output. We select 4004 instances uniformly distributed over the 14 patterns, which are split into train:dev:test -- 2576:630:798.

\subsection{Multiple Choice Problems: Blackbird Language Matrices}
\label{sec:BLMs}

Blackbird Language Matrices (BLMs) \citep{merlo2023} are language versions of the visual Raven Progressive Matrices (RPMs). Like the RPMs, they are multiple-choice problems. The input is a sequence of 7 sentences built using specific rules, and the correct answer fits within the sequence defined by these rules. The incorrect options are built by corrupting some of the underlying generating rules of the input sentence sequence. Solving the problem requires identifying the entities (the chunks), their relevant  attributes (their morphological or semantic properties) and their connecting operators.  

We use two BLM datasets: (i) BLM-AgrF -- subject verb agreement in French \citep{an-etal-2023-blm}, and (ii) BLM-s/lE -- the spray-load verb alternations in English\footnote{Agent-Location-Theme (ALT) -- Agent-Theme-Location (ATL)} \citep{samo-etal-2023}. The structure of these datasets -- in terms of the sentence chunks and sequence structure -- is shown in Figure \ref{fig:BLM_structure}.

\paragraph{Datasets statistics}
Table \ref{tab:data} shows the datasets statistics. Each set is split 90:10 into train:test subsets, and then we randomly sample 2000 instances as train data. 20\% of the train data is used for development. Types I, II, III correspond to different amounts of lexical variation within an instance.

\begin{table}[h]
  \small
    \begin{tabular}{l|r|rr}
     & Subj.-verb agr. & \multicolumn{2}{c}{Verb alternations} \\ 
     &                 & ALT-ATL & ATL-ALT \\ \hline
    Type I  & 2000:252  &  2000:375 & 2000:375 \\
    Type II & 2000:4866 &  2000:1500 & 2000:1500 \\
    Type III & 2000:4869 &  2000:1500 & 2000:1500 \\ \hline
    \end{tabular}
    \caption{Train:Test statistics for the two BLM problems. }
    \label{tab:data}
\end{table}

\noindent To solve a BLM instance, the system processes the input sentence sequence and outputs a sentence representation that will be compared to the representation of the sentences in the answer set. The candidate answer closest to the generated sentence representation will be considered the correct one.

We run the experiments on the BLMs for agreement and on the verb alternation BLMs.
While the information necessary to solve the agreement task is more structural, solving
the verb alternation task requires not only structural information on chunks, but also
semantic information, as syntactically similar chunks play different roles in a sentence.

\section{Experiments}

We present a progression of experiments.
\begin{enumerate}
    \item Using the dataset of sentences with known chunk structure, we test whether a sparse variational encoder-decoder system can distill information about the chunk structure of a sentence from its embedding.
    \item We analyze the sparse model, and trace the information from the latent layer back to the sentence embedding to understand where in the sentence embeddings these differences are encoded.
    \item We combine the sparsified variational encoder-decoder with another VAE-like layer to solve the BLM tasks, and test whether the latent layer sentence encodings maintain information useful for the tasks.
    \end{enumerate}

All experiments use Electra \citep{clark-etal-2019-bert}\footnote{Electra pretrained model: google/electra-base-discriminator}. We use as sentence representations the embedding of the [CLS] token, reshaped as a two dimensional array with shape 32x24.

The experiments are analyzed through the output of the system, in terms of average F1 score over three runs. For the investigations of the sentence embeddings, we also analyze the compressed vectors on the latent layer, to determine whether chunk patterns are encoded in these vectors. If these vectors cluster by the chunk pattern of the corresponding sentences it will indicate that sentence chunk patterns were indeed detected and are encoded differently in the latent layer. 

\subsection{Sparsification}
\label{sec:sparsification}

\citet{nastase2024identifiable}\/ have shown that sentence embeddings contain information about the chunk structure and their properties using an encoder-decoder architecture that compresses the relevant information into a small latent layer. They build on \citet{nastase-merlo-2023-grammatical} who show that reshaping a sentence embedding from the commonly used one-dimensional array to a two-dimensional representation allows grammatical information to become more readily accessible. 

\begin{figure}[h]
    \centering
    \includegraphics[width=0.48\textwidth]{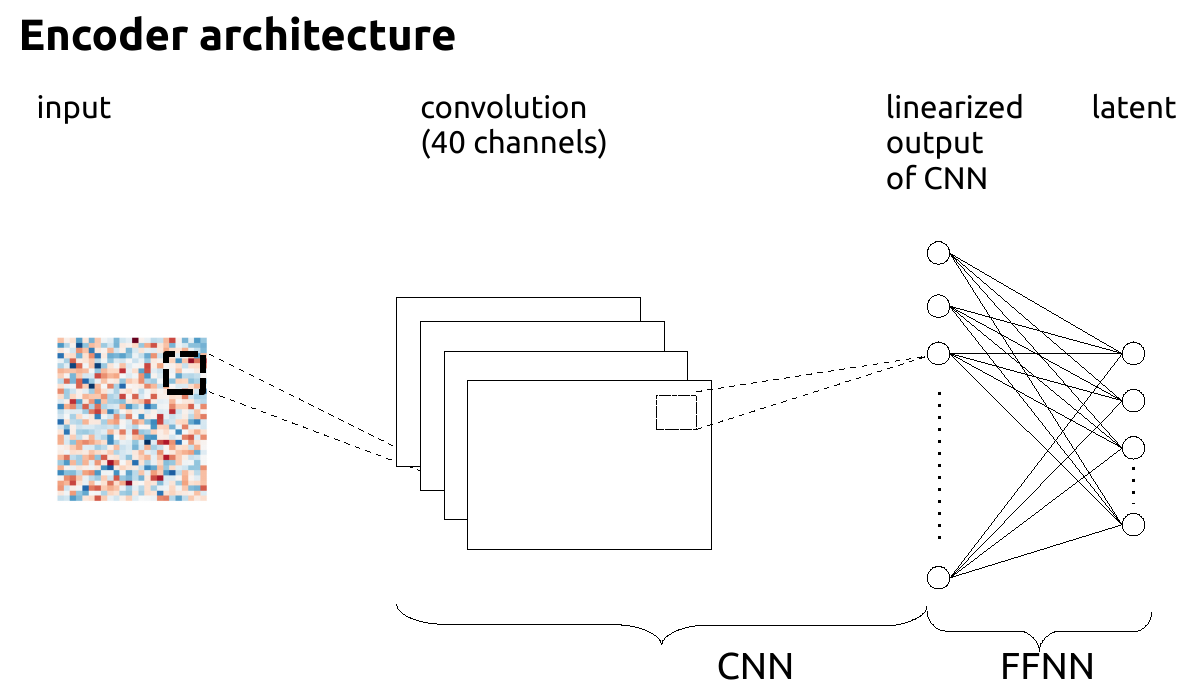}
    \caption{Details of the encoder architecture}
    \label{fig:encoder}
\end{figure}

We adopt the system of \cite{nastase2024identifiable}, with the same architecture (including number of CNN channels and kernel size), and sparsify it, to determine whether specific information can be localized in sentence embeddings. The encoder of the system consists of a CNN layer followed by a FFNN, that compresses the information into a latent layer, as illustrated in Figure~\ref{fig:encoder}.

The CNN layer in the encoder detects a different pattern in the sentence representation on each of its 40 channels. The linear layer compresses the linearized output of the CNN into a very small latent layer (length 5). A vector is sampled from this, and then decoded into a sentence representation using a decoder which is a mirror of the encoder. 

An instance consists of an input sentence $s$, and 7 output sentences, only one of which has the same chunk structure as the input and is considered the correct one (section \ref{sec:sent_descr}). The aim is to guide the system to capture information about the chunk structure of the sentences in the latent layer, by using a max-margin loss function  that assigns a higher score to the correct option relative to the others. Formally, if $e_s$ is the embedding of the input sentence $s$, $\hat{e}_s$ is the embedding output by the decoder, $e_c$ is the embedding of the correct option and $e_i,$ $i=1,6$ are the embeddings of the other options, and $mm$ is the maxmargin function, then:

    \vspace{2mm}
\noindent {\small \[loss(s) = 
mm(\hat{e}_s,e_c,\{e_i|i=1,6\}) + KL(z_s||\mathcal{N}(0,1))\]}

\noindent {\small $mm(\hat{e}_s,e_c,e_i)$ = 

$max(0, 1-score(\hat{e}_s, e_c) + \sum_{i=1}^{6} score(\hat{e}_s, e_i) / 6)$}
    \vspace{2mm}
    
\begin{figure}[h]
    \centering
   \includegraphics[width=0.4\textwidth]{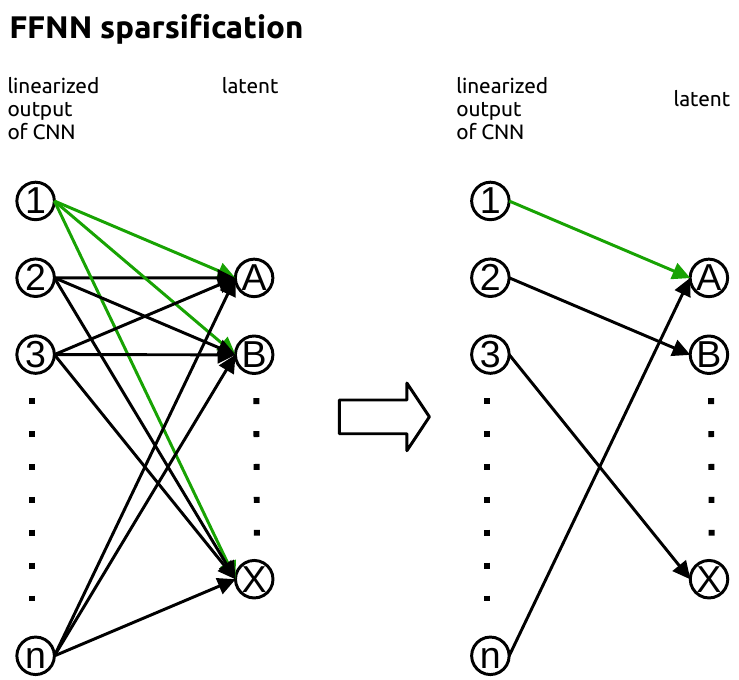}
    \caption{Separating linguistic signals by masking the one-layer FFNN}
    \label{fig:separation}
\end{figure}

We want to sparsify this network in a targeted way: we enforce that each output unit from the CNN layer will contribute to only one unit in the latent layer. Figure \ref{fig:separation} illustrates the process.

To enforce this behaviour, we use an approach inspired from sparsification \cite{savarese2020-sparsification} and subnetworking \cite{lepori2023break}. Instead of considering the output of the CNN as the input layer of a linear network, we make each CNN output unit the input of a separate linear network, connected to the latent layer. We apply a mask $m$ to the weights $W$ of this network, and compute a masked weight matrix $W_m = W \times softmax(M/\tau)$, where $\tau$ is a temperature parameter used to push the softmax function towards a one-hot vector.

We use a kernel 15x15\footnote{We adopt the size of the kernel from previous work.} and equal stride (15x15) to have a very clear separation of the information flow from the sentence embedding to the latent layer. This will ensure our sparsification desideratum, and the learned network will have a particular configuration: if $N_{CNN}$ is the set of output nodes from the CNN, and $N_L$ are the nodes on the latent layer, then the sets of CNN output nodes connected to each of the latent units are disjunct:

    \vspace{2mm}
\noindent{\small $\forall n_l \in N_L,  S_{CNN}^l = \{n_c \in N_{CNN} | W_m(n_l, n_c) > 0\} $}

    \vspace{2mm}
    $\mbox{and if } i \ne j \mbox{ then } S_{CNN}^i \cap S_{CNN}^j = \varnothing$

\paragraph{Sparsification results}

\begin{figure}[h]
    \centering
    \includegraphics[width=0.45\textwidth]{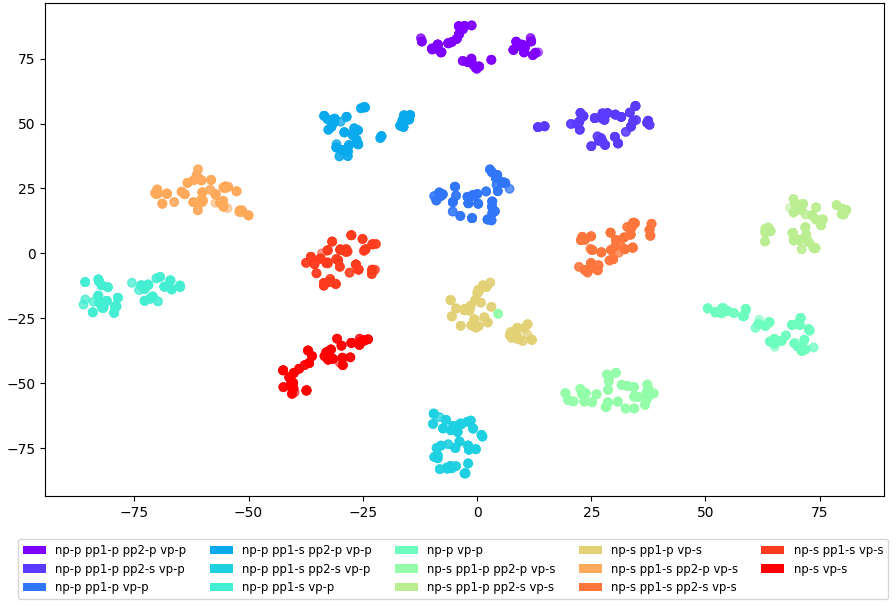}
    \caption{TSNE projection of the latent layer for encoder-decoder with full network connections.}
    \includegraphics[width=0.45\textwidth]{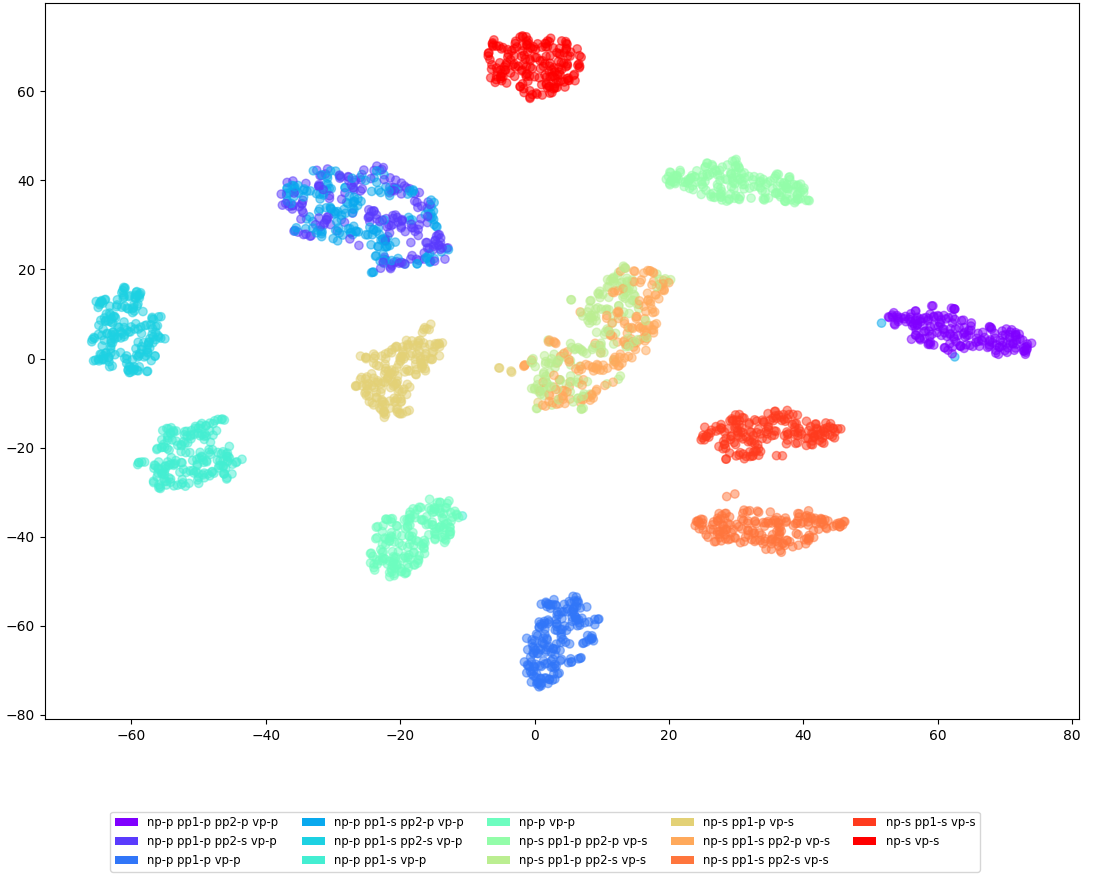}    
    \caption{TSNE projection of the latent layer for sparsified encoder-decoder.}
    \label{fig:latent_plots}
\end{figure}

Despite the fact that this type of sparsification is very harsh, and channels the information from the sentence embedding into very few paths on the way to the latent layer, the results in terms of average F1-score/standard deviation over three runs without  0.997 (0.0035) and with sparsification 0.977 (0.0095) are close. While this difference is rather small, we notice a bigger difference in the latent layer. Figure \ref{fig:latent_plots} shows the TSNE projections of the latent layers.
As can be seen,  while the full network shows a very clear and crisp separation of latents that encode different chunk patterns -- with a 0.9928/0.0101 F1 macro-average/standard deviation -- when sparsifying the information is slightly less crisp in the 2D TSNE projection, but still high F1 macro-average/standard deviation (0.9886/0.0038) 

\subsection{Localizing linguistic information in sentence embeddings}

We approach the isolation of linguistic information with the following intuition: on each channel, the CNN discovers different patterns in various regions of the sentences. Some combination of these patterns -- i.e. some combinations of signals from the CNN output -- encode specific properties of the sentences. These signals eventually reach the latent layer. Previous experiments have shown that this latent layer contains information about chunks and their properties. Working backwards from the latent layer to the sentence embedding -- through the CNN output layer, the different channels and sentence embedding regions --  helps us trace back where the biggest changes are when the input sentences have different properties.

To verify whether specific linguistic information, like different number of chunks, or different chunk properties, is encoded in different regions of the sentence embeddings, we analyse the distribution of values in each network node in the encoder, namely the CNN output nodes $N_{CNN}$ and the latent nodes $N_L$. 

We denote $S_p$ the set of input sentences that share the same chunk pattern $p$ (for instance, $p$ = "NP-s VP-s"). We pass their sentence embeddings through the learned encoder, and gather the values in each CNN output node:

    $V_{CNN}^p = \{V_{CNN}^p(n_c) | n_c \in N_{CNN}\}$

    $V_{CNN}^p(n_c) = \{ val_{n_c}(s) | s \in S_p \}$

\noindent and $val_{n_c}(s)$ is the value in the CNN output node $n_c$ when the input is the embedding of sentence $s$.

To check for differences in how sentence with different patterns are encoded, we will look at sets of sentences $S_{p_1}$ and $S_{p_2}$ where $p_1$ and $p_2$ are patterns that differ minimally. We consider three such minimal differences:
    \begin{description}
        \item[length] one pattern has an extra (or one less) chunk than the other but are otherwise identical ({\it np-s vp-s} vs. {\it np-s \ul{pp1-s} vp-s}),
        \item[grammatical number] the two patterns have the same number of chunks, but one (and only one) chunk has a different grammatical number than the other ({\it np-s pp1-\ul{s} vp-s} vs. {\it np-s pp1-\ul{p} vp-s}),
        \item[subject-verb number alternation] the two patterns are identical except in the grammatical number of the subject and verb ({\it np-\ul{s} pp1-s vp-\ul{s}} vs. {\it np-\ul{p} pp1-s vp-\ul{p}}).
    \end{description}

\begin{figure*}
    \includegraphics[width=0.33\textwidth]{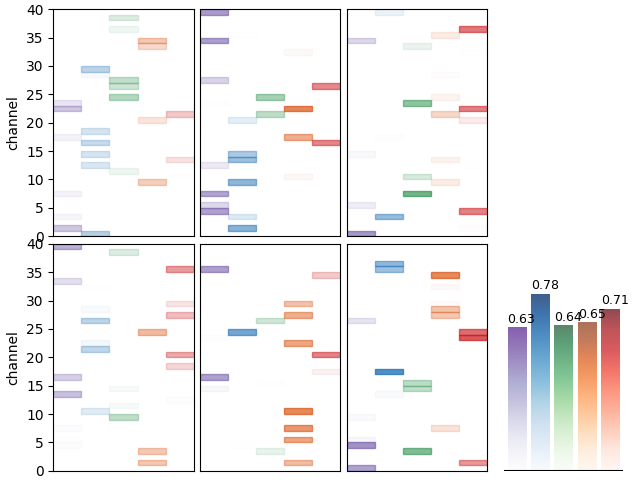}
    \includegraphics[width=0.33\textwidth]{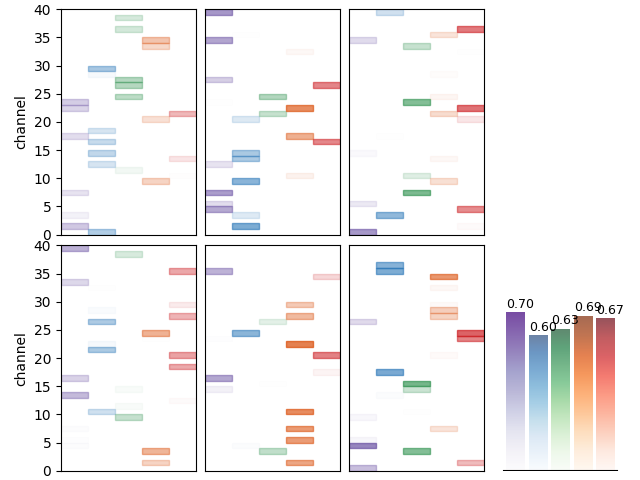}
    \includegraphics[width=0.33\textwidth]{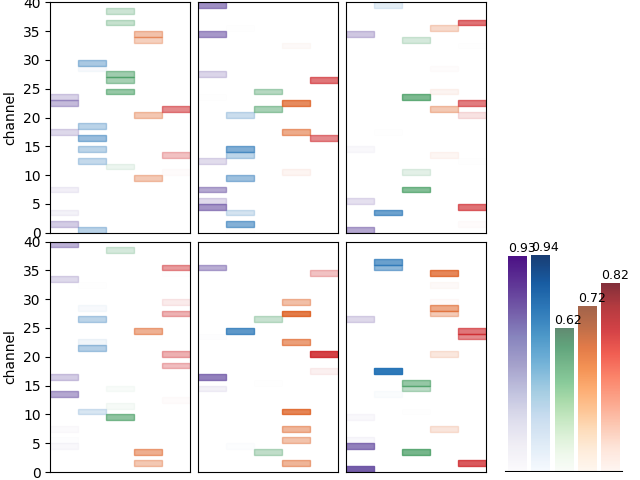}
    \caption{Average cosine distance between value distributions in each CNN output node (i.e. each node corresponding to the application of the kernel from each channel on the sentence embeddings, according to the kernel size and stride) for sets of sentences with minimally different patters: (left) patterns differ in only one grammatical number attribute for one chunk, (middle) patterns differ only in length, (right) patterns differ only in the number of the subject and verb. Each panel corresponds to one region of the sentence embedding the size of the kernel. The y-axis represents the channels of the CNN. The x-axis represents the latent units in different colours (the stronger the color, the higher the value, max = 1), and the pairs of compared patterns represented as adjacent rectangles. }
    \label{fig:pattern_comparison_compressed}
\end{figure*}

To compare how chunk information is encoded in sentences that have different patterns $p_1$ and $p_2$, we compare the sets of values in each CNN output node $n_c$: $V_{CNN}^{p_1}(n_c)$ and $V_{CNN}^{p_2}(n_c)$ . If these value distributions are very different, this is an indication that the area of a sentence embedding where the signal to $n_c$ is coming from is involved in encoding the type of information that is different between $p_1$ and $p_2$. 

We perform this analysis in two steps: (i) a filtering step that eliminates from the analysis the CNN output nodes that do not encode differences in behaviour between patterns, and (ii) a quantification of the differences in the values in the node for different patterns. 

The filtering step is performed using a two-sample Kolmogorov-Smirnov test \cite{Hodges1958TheSP},\footnote{We use the ks\_2samp test in the scipy Python package} which provides information whether two samples come from the same distribution. As we are interested in the CNN output nodes where the value distributions are different when the inputs are sentences with different patterns, we will filter out from the analysis the nodes $n_c$ where the sets of values $V_{CNN}^{p}(n_c)$ come from the same distribution for all patterns $p$ represented in the data. 

For the remaining CNN output nodes, we project the value distributions onto the same set of bins, and then quantify the difference using cosine distance. Specifically, we determine the range of values for $V_{CNN}^p$ for all patterns $p$ -- $min_{V_{CNN}}, max_{V_{CNN}}$, and split it into 100 bins. For each CNN output node $n_c$ and pattern $p$ we make a value distribution vector $v^p_{n_c}$ from the node's set of values $V_{CNN}^p(n_c)$, w.r.t. the 100 bins.

We then compute a score for every pair of minimally different patterns $p_1, p_2$ for each node $n_c$ as the cosine distance:

$score_{n_c}(p_1,p_2) = 1-cos(v^{p_1}_{n_c}, v^{p_2}_{n_c})$

This score quantifies how different a region of the sentence embedding is when encoding sentences with different chunk patterns.

\begin{figure*}
    \includegraphics[width=0.31\textwidth]{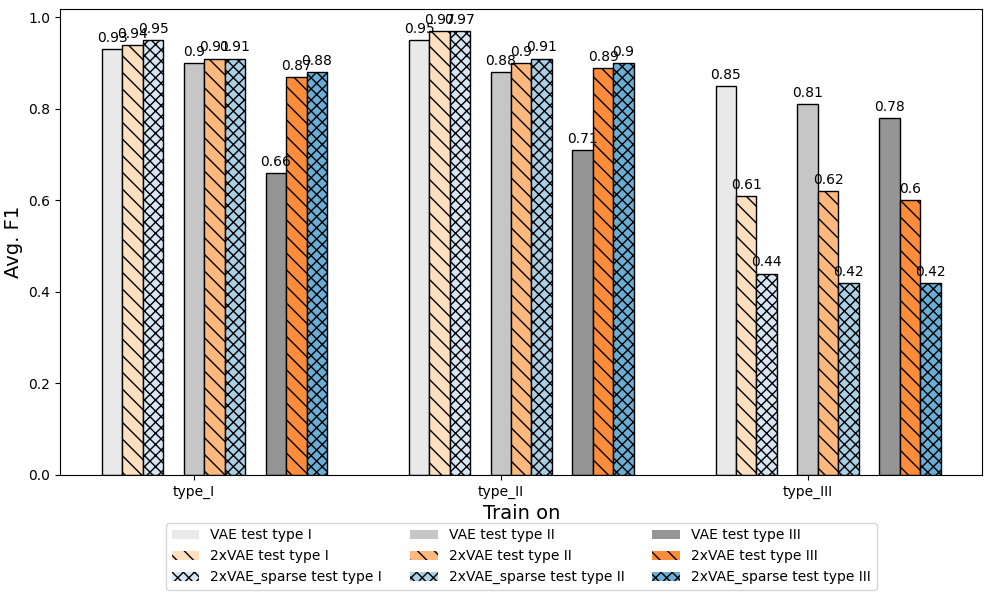} \hfill \includegraphics[width=0.30\textwidth]{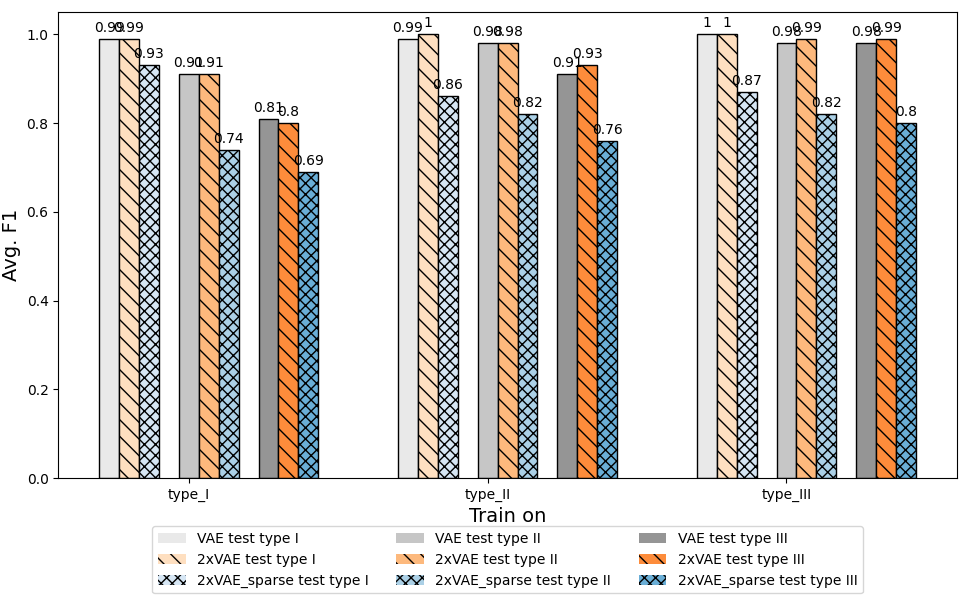} \hfill \includegraphics[width=0.37\textwidth]{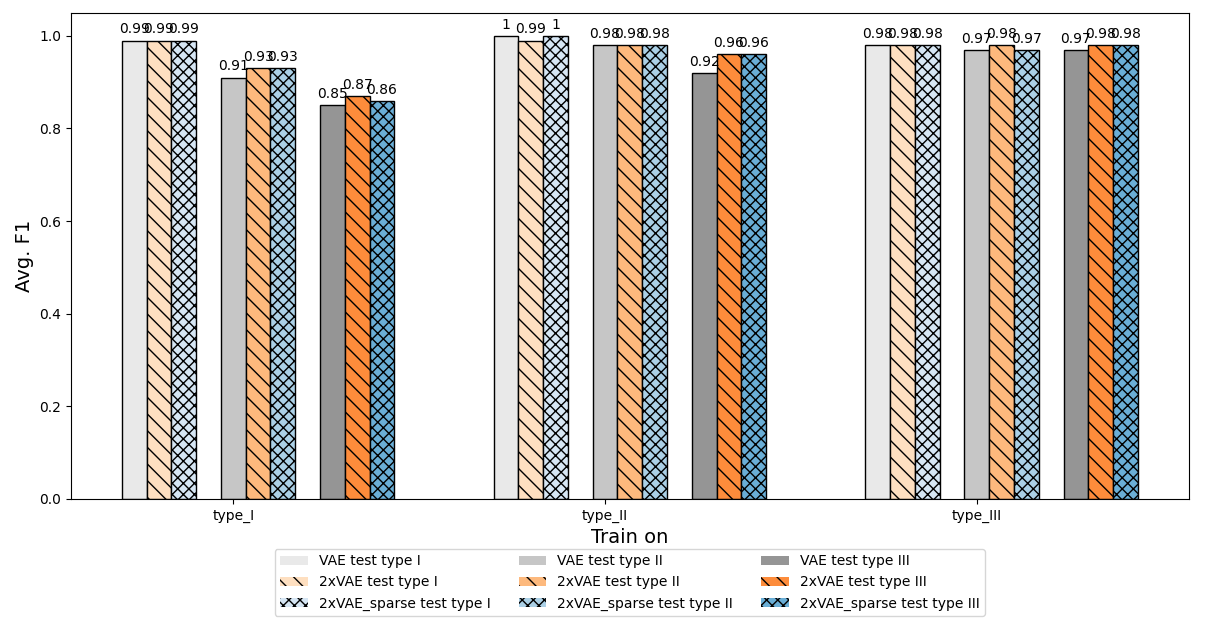}
    \caption{Results in term of average F1 scores over 3 runs, for the BLM agreement (1) and verb alternations ALT-ATL (2) and ATL-ALT (3) }
    \label{fig:blm_results}
\end{figure*}

\paragraph{Localization results}

A first clue that information related to chunk patterns in a sentence is localized is the fact that the filtering step using the two-sample Kolmogorov-Smirnov test leads to the removal of 83 out of 240 (34\%) CNN output nodes. 

For the remaining nodes where differences in value distributions between different sentence patterns exist, we compute the cosine distance between pairs of minimally different patterns with respect to grammatical number, length and subject-verb number alternations. Figure \ref{fig:pattern_comparison_compressed}  shows the differences in value distributions in each CNN output nodes from each channel -- channels are reprezented on the y-axis, and the 5 latent units on the x-axis in different colours. A stronger colour indicates a stronger effect. More detailed plots are included in Figure \ref{fig:pattern_comparison_detailed} in the appendix. 

\begin{wrapfigure}{r}{0.14\textwidth}
\includegraphics[width=0.14\textwidth]{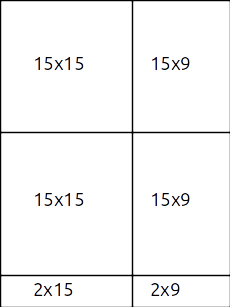}
\end{wrapfigure}

These plots indicate that there are few channel-sentence region combinations that encode differences in chunk structure in the input sentences. While in the figure the sentence areas are illustrated with equal sizes, the regions are presented transposed for space considerations, and they have the shapes shown in the adjacent figure. The chunks and the chunk information seems to be encoded in the bottom part of the sentence embedding, and much of it in the bottom 2x24 area.


\subsection{BLM tasks}

\begin{figure}
    \centering
    \includegraphics[width=0.45\textwidth]{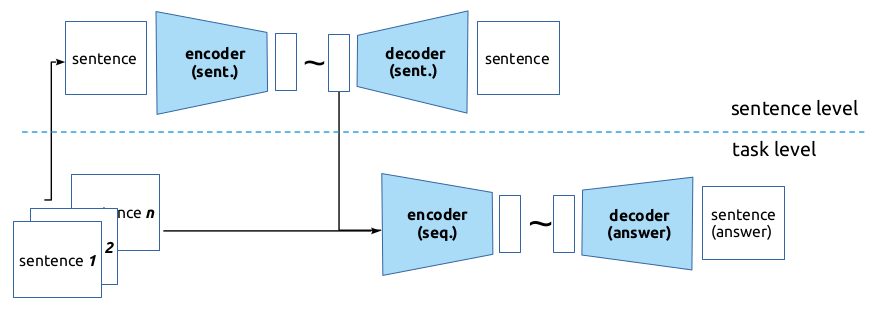}
    \caption{A two-level variational encoder-decoder: the top level compresses the sentence embeddings into a latent layer, and the bottom level uses the compressed sentence representations to solve the BLM tasks.}
    \label{fig:2level_VAE}
\end{figure}

To further test whether task specific information is robust to sparsification, we use the two-level variational encoder-decoder depicted in Figure \ref{fig:2level_VAE}.

An instance for a BLM task consists of a tuple, comprising a sequence of sentences $S = \{s_i |i = 1, 7\}$ as input, and an answer set with one correct answer $a_c$, and several incorrect answers $a_{err}$. The sentence level of the 2-level encoder-decoder compresses the sentence embeddings of each of the sentences in the input sequence into a small latent vector. The sampled latent representations are then used as the representations of the sentences in the
input sequence. This sequence representation is passed as input to the BLM-level encoder-decoder, it is compressed into a new latent layer, and the sampled vector is then decoded into a sentence representation that  best matches the representation of the correct answer.

\paragraph{BLM task results}

We evaluate the performance of the sparsified 2-level VAE on the BLM tasks. Only the first level of the VAE, the one processing individual sentences, is sparsified as described in section \ref{sec:sparsification}. Figure \ref{fig:blm_results} shows the performance of three system variations: (i) a one-level VAE that processes the input sequence of sentences and produces a sentence representation, (ii) the two-level VAE described in more detail in \citep{nastase2024identifiable}, (iii) the sparsified version of the sentence level VAE in the two-level VAE. As in the previous experiments, sparsification does not cause harsh drops in performance for either of the two BLM tasks. The reason for this is the same reason we chose this particular data for experiments: solving the task relies on the system having information about the chunks in the sentence, and their properties. As long as that type of information is preserved, the tasks can be solved successfully. 

We note two main changes however. In the agreement task, the sparsified system registers a drop in performance when trained on maximally lexically different data (type III). The two-level system without sparsification also registers such a drop in comparison with the baseline one-level encoder decoder. Both these effects may be due to the ambiguous supervision signal at the sentence level of the system: while using type I and type II data with little lexical variation, it is easier for the system to focus on structural differences between the correct and incorrect output options. When using type III data with much lexical variation, it is not clear for the system what is the relevant dimension of difference between the output options. 

In the verb alternation task, previous results on predicting the Agent-Theme-Location or the Agent-Location-Theme alternation produced very similar results. This is not the case here, but understanding why this happens requires additional analysis.

\section{Conclusions}

Our aim was to understand how information is encoded in sentence embedding, given that previous work has shown that various types of linguistic information is encoded in a model's layers and parameters. We investigated this question using a dataset of sentences with specific chunk structure, and two multiple-choice problems that require information about sentence chunks and their properties to be solved successfully. We have shown that using a sparsified encoder-decoder system, the sentence representations can be compressed into a latent layer that encodes chunk structure properties. We then traced back the signal from the latent layer to the sentence embedding, to detect which areas of the sentence embeddings change the most when comparing sentences with different chunk patterns. This analysis shows that such information is captured by a small number of channel-sentence area combinations. Further experiments with the two multiple-choice tasks have confirmed that chunk information and their grammatical properties (for the agreement BLM) and chunk information and their semantic role properties (for the verb alternation BLM) are captured by the sparsified sentence compression level. We envision further analyses to see where the differences between chunk patterns that have different semantic roles are encoded, and get closer to decoding the sentence embeddings.

\section{Limitations}

We have explored sentence embeddings using an artificially constructed dataset with simple chunk structure. To check how this kind of information is localized, we started from a previously developed system that showed high performance in distinguishing the patterns of interest. We have not changed the system's parameters (such as the kernel size of the CNNs), and have not performed additional parameter search to narrow down the locations to smaller regions. We plan to address sentence complexity issues and parameters for narrower localization of information in future work.

\paragraph{Acknowledgments}

We gratefully acknowledge the partial support of this work by the Swiss National Science Foundation, through grant SNF Advanced grant  TMAG-1\_209426 to PM.

\bibliographystyle{acl_natbib}
\bibliography{anthology_since2015,custom}

\appendix
\include{appendix}

\end{document}